\documentclass[11pt,a4paper]{article}

\usepackage[hyperref]{eacl2021}

\usepackage{times}
\usepackage{latexsym}

\usepackage{microtype}

\usepackage{booktabs}
\usepackage{graphicx}
\usepackage{subfigure}
\usepackage{amsmath}
\usepackage{amssymb}
\usepackage{fontawesome}
\usepackage{hyperref}
\usepackage{url}
\usepackage{enumitem}
\usepackage{comment}
\usepackage{float}
\usepackage{placeins}
\usepackage{multibib}
\usepackage[colorinlistoftodos]{todonotes}

\newcommand{\multirowcell}[1]{\begin{tabular}[c]{@{}c@{}}#1\end{tabular}}

\aclfinalcopy 
\def\aclpaperid{690} 

\title{Active Learning for Sequence Tagging with Deep Pre-trained Models and Bayesian Uncertainty Estimates}

\author{Artem Shelmanov$^1$, Dmitri Puzyrev$^{1,2,3}$, Lyubov Kupriyanova$^1$,  Denis Belyakov$^2$, \\ \textbf{Daniil Larionov$^{2,3,4}$,  Nikita Khromov$^1$, Olga Kozlova$^3$, Ekaterina Artemova$^2$,}\\ \textbf{Dmitry V. Dylov$^1$ and Alexander Panchenko$^1$}  \\
  $^1$Skolkovo Institute of Science and Technology, Russia \\
  $^2$HSE University, Russia \\
  $^3$Mobile TeleSystems (MTS), Russia \\
  $^4$Federal Research Center ``Computer Science and Control'' of Russian Academy of Sciences \\
 \href{mailto:artemshelmanov@gmail.com}{a.shelmanov@skoltech.ru}\\ 

\\}

\date{}

\begin{document}
\maketitle

\begin{abstract}

Annotating training data for sequence tagging of texts is usually very time-consuming. Recent advances in transfer learning for natural language processing in conjunction with active learning open the possibility to significantly reduce the necessary annotation budget. We are the first to thoroughly investigate this powerful combination for the sequence tagging task. We conduct an extensive empirical study of various Bayesian uncertainty estimation methods and Monte Carlo dropout options for deep pre-trained models in the active learning framework and find the best combinations for different types of models. Besides, we also demonstrate that to acquire instances during active learning, a full-size Transformer can be substituted with a distilled version, which yields better computational performance and reduces obstacles for applying deep active learning in practice.
\end{abstract}


\section{Introduction}
\label{intro}

In many natural language processing (NLP) tasks, such as named entity recognition (NER), obtaining gold standard labels for constructing the training dataset can be very time and labor-consuming. It makes the annotation process expensive and limits the application of supervised machine learning models. This is especially the case in such domains as biomedical or scientific text processing, where  crowdsourcing is either difficult or prohibitively expensive. In these domains, highly-qualified experts are needed to annotate data correctly, which dramatically increases the annotation cost.

Active Learning (AL) is a technique that can help to reduce the amount of annotation required to train a good model by multiple times \cite{settles-craven-2008-analysis,Settles2009}. Opposite to exhaustive and redundant manual annotation of the entire corpus, AL drives the annotation process to focus the expensive human expert time only on the most informative objects, which contributes to a substantial increase in the model quality.

AL is an iterative process that starts from a small number of labeled seeding instances. In each iteration, an \textit{acquisition} model is trained on the currently annotated dataset and is applied to the large pool of unannotated objects. The model predictions are used by the AL query strategy to sample the informative objects, which are then further demonstrated to the expert annotators. When the annotators provide labels for these objects, the next iteration begins. The collected data can be used for training a final \textit{successor} model that is used in a target application.

During AL, acquisition models have to be trained on very small amounts of the labeled data, especially during the early iterations. Recently, this problem has been tackled by transfer learning with deep pre-trained models: ELMo \cite{peters2018deep}, BERT \cite{Devlin2019BERTPO}, ELECTRA \cite{electra}, and others. Pre-trained on a large amount of unlabeled data, they are capable of demonstrating remarkable performance when only hundreds or even dozens of labeled training instances are available. This trait suits the AL framework but poses the question about the usefulness of the biased sampling provided by the AL query strategies.  

In this work, we investigate AL with the aforementioned deep pre-trained models and compare the results of this combination to the outcome of the models that do not take advantage of deep pre-training. The main contributions of this paper are the following: 
\begin{itemize}
    \item We are the first to thoroughly investigate deep pre-trained models in the AL setting for sequence tagging of natural language texts on the widely-used benchmarks in this area. 
    \item We conduct an extensive empirical study of various AL query strategies, including Bayesian uncertainty estimation methods with multiple Monte Carlo (MC) dropout variants \cite{gal2016dropout,gal2017deep}. We find the best combinations of uncertainty estimates and dropout options for different types of deep pre-trained models.
    \item We show that to acquire instances during AL,
    a full-size Transformer can be substituted with a distilled version, which yields better computational performance and reduces obstacles for applying deep AL in practice.
\end{itemize}

The remainder paper is structured as follows. Section \ref{relwork} covers relevant works on AL for sequence tagging. In Section \ref{sec:st_models}, we describe the sequence tagging models. Section \ref{sec:al_qs} describes the AL strategies used in the experiments. In Section \ref{sec:exps}, we discuss the experimental setup and present the evaluation results. Finally, Section \ref{sec:conclusion} concludes the paper.


\section{Related Work}
\label{relwork}
                        
AL for sequence tagging with classical machine learning algorithms and a feature-engineering approach  has a long research history, e.g. \cite{settles-craven-2008-analysis,Settles2009,marcheggiani2014experimental}. More recently, AL in conjunction with deep learning has received much attention.

In one of the first works on this topic, \newcite{shen2018} note that practical deep learning models that can be used in AL should be computational efficient both for training and inference to reduce the delays in the annotators' work. They propose a CNN-CNN-LSTM architecture with convolutional character and word encoders and an LSTM tag decoder, which is a faster alternative to the widely adopted LSTM-CRF architecture \cite{lample2016} with comparable quality. They also reveal disadvantages of the standard query strategy -- least confident (LC), and propose a modification, namely Maximum Normalized Log-Probability (MNLP). \newcite{Siddhant2018DeepBA} experiment with Bayesian uncertainty estimates. They use CNN-CNN-LSTM and CNN-BiLSTM-CRF \cite{ma2016end} networks and two methods for calculating the uncertainty estimates: Bayes-by-Backprop \cite{blundell2015weight} and the MC dropout \cite{gal2016dropout}. The experiments show that the variation ratio \cite{freeman1965elementary} has substantial improvements over MNLP. In contrast to them, we additionally experiment with the Bayesian active learning by disagreement (BALD) query strategy proposed by \newcite{houlsby2011bayesian} and perform a comparison with variation ratio. 

There is a series of works that tackle AL with a trainable policy model that serves as a query strategy. For this purpose, imitation learning is used in \newcite{liu2018learning,vu2019learning,brantley-etal-2020-active}, while in \cite{fang2017learning}, the authors use deep reinforcement learning. Although the proposed solutions are shown to outperform other heuristic algorithms with comparably weak models (basic CRF or BERT without fine-tuning) in experiments with a small number of AL iterations, they can be not very practical due to the high computational costs of collecting training data for policy models. Other notable works on deep active learning include \cite{erdmann2019}, which proposes an AL algorithm based on a bootstrapping approach \cite{jones1999bootstrapping} and  \cite{lowell2019practical}, which concerns the problem of the mismatch between a model used to construct a training dataset via AL (acquisition model) and a final model that is trained on it (successor model). 

Deep pre-trained models are evaluated in the AL setting for NER by \newcite{shelmanov2019}. However, they perform the evaluation only on the specific biomedical datasets and do not consider the Bayesian query strategies. \newcite{eindoretal2020active} conduct an empirical study of AL with pre-trained BERT but only on the text classification task. \newcite{brantley-etal-2020-active} use pre-trained BERT in experiments with NER, but they do not fine-tune it, which results in suboptimal performance. In this work, we try to fill the gap by evaluating deep pre-trained models: ELMo and various Transformers, in the AL setting with practical query strategies, 
including Bayesian, and on the widely-used benchmarks in this area.


\section{Sequence Tagging Models}
\label{sec:st_models}

We use a tagger based on the Conditional Random Field model \cite{Lafferty2001}, two BiLSTM-CRF taggers \cite{lample2016} with different word representation models, and taggers based on state-of-the-art Transformer models.

\subsection{Conditional Random Field}

As a baseline for comparison, we use a feature-based linear-chain Conditional Random Field (CRF) model \cite{Lafferty2001}. It is trained to maximize the conditional log-likelihood of entire tag sequences. The inference is performed using the Viterbi decoding algorithm, which maximizes the joint probability of tags of all tokens in a sequence. The features used for the CRF model are presented in Appendix \ref{appendixA}.

\subsection{BiLSTM-CRF}

This model encodes embedded input tokens via a bidirectional long short term memory neural network (LSTM) \cite{Hochreiter1997}. BiLSTM processes sequences in two passes: from left-to-right and from right-to-left producing a contextualized token vector in each pass. These vectors are concatenated and are used as features in a CRF layer that performs the final scoring of tags.

We experiment with two versions of the BiLSTM-CRF model. The first one uses GloVe \cite{pennington2014glove} word embeddings pre-trained on English Wikipedia and the 5-th edition of the Gigaword corpus, and a convolutional character encoder \cite{ma2016end}, which helps to deal with out-of-vocabulary words. As in \cite{chiu2016named}, the model additionally leverages the basic capitalization features, which has been shown to be useful for achieving good performance with this model. We will refer to it as CNN-BiLSTM-CRF. We consider this model as another baseline that does not exploit deep pre-training. The second version of the BiLSTM-CRF model uses pre-trained medium-size ELMo \cite{peters2018deep} to produce contextualized word representations. ELMo is a BiLSTM language model enhanced with a CNN character encoder. This model does not rely on feature-engineering at all. We will refer to it as ELMo-BiLSTM-CRF.

\subsection{Transformer-based Taggers}

We perform AL experiments with state-of-the-art pre-trained Transformers: BERT \cite{Devlin2019BERTPO}, DistilBERT \cite{sanh2019distilbert}, and ELECTRA \cite{electra}. The sequence tagger, in this case, consists of a Transformer ``body'' and a decoding classifier with one linear layer. Unlike BiLSTM that encodes text sequentially, these Transformers are designed to process the whole token sequence in parallel with the help of the self-attention mechanism \cite{vaswani2017attention}. This mechanism is bi-directional since it encodes each token on multiple neural network layers taking into account all other token representations in a sequence. These models are usually faster than the recurrent counterparts and show remarkable performance on many downstream tasks \cite{li2020survey}.

BERT is a masked language model (MLM). Its main learning objective is to restore randomly masked tokens, so it can be considered as a variant of a denoising autoencoder. Although this objective makes the model to learn many aspects of natural languages \cite{tenney-etal-2019-bert,rogers2020primer}, it has multiple drawbacks, including the fact that training is performed only using a small subset of masked tokens. ELECTRA has almost the same architecture as BERT but utilizes a novel pre-training objective, called replaced token detection (RTD), which is inspired by generative adversarial networks. In this task, the model has to determine what tokens in the input are corrupted by a separate generative model, in particular, a smaller version of BERT. Therefore, the model has to classify all tokens in the sequence, which increases training efficiency compared to BERT, and the RTD task is usually harder than MLM, which makes the model learn a better understanding of a language \cite{electra}. 

DistilBERT is a widely-used compact version of BERT obtained via a distillation procedure \cite{hinton2015distilling}. The main advantages of this model are the smaller memory consumption and the higher fine-tuning and inference speed achieved by sacrificing the quality. We note that good computational performance is a must for the practical applicability of AL. Delays in the interactions between a human annotator and an AL system can be expensive. Therefore, although DistilBERT is inferior compared to other Transformers in terms of quality, it is a computationally cheaper alternative for acquiring training instances during AL that could be used for fine-tuning bigger counterparts. \newcite{lowell2019practical} showed that a mismatch between an acquisition model and a successor model (the model that is trained on the annotated data for the final application) could eliminate the benefits of AL. The similarity in the architectures and the shared knowledge between the smaller distilled Transformer and its ancestor potentially can help to alleviate this problem and deliver an AL solution that is both effective and practical.


\section{Active Learning Query Strategies}
\label{sec:al_qs}

We experiment with four query strategies for the selection of training instances during AL. 

\textit{\textbf{Random sampling}} is the simplest query strategy possible: we just randomly select instances from the unlabeled pool for annotation. In this case, there is no active learning at all.

Uncertainty sampling \cite{lewis1994sequential} methods select instances according to some probabilistic criteria, which indicates how uncertain the model is about the label that was given to the instance. The baseline method is Least Confident (LC): the samples are sorted in the ascending order of probabilities of the most likely tag sequence. Let $y_i$ be a tag of a token $i$ that can take one class $c$ out of $C$ values, let $x_j$ be a representation of a token $j$ in an input sequence of length $n$. Then the LC score can be formulated as follows:
\[
LC = 1 - \max _{y_{1}, \ldots, y_{n}} \mathbb{P}\left[y_{1}, \ldots, y_{n} |\left\{\mathbf{x}_{j}\right\}\right]
\]

This score favors longer sentences since long sentences usually have a lower probability. Maximization of probability is equivalent to maximizing the sum of log-probabilities:
\[
\begin{aligned}
\max _{y_{1}, \ldots, y_{n}} \mathbb{P}\left[ y_{1}, \ldots, y_{n} |\{\mathbf{x}_{j}\} \right] \Leftrightarrow \\
 \Leftrightarrow \max _{y_{1}, \ldots, y_{n}} \sum_{i}^{n} \log \mathbb{P}\left[y_{i} | \{y_j\}\setminus y_i,\left\{\mathbf{x}_{j}\right\}\right]
\end{aligned}
\]

To make LC less biased towards longer sentences, \newcite{shen2018} propose a normalization of the log-probability sum. They call the method \textit{\textbf{Maximum Normalized Log-Probability (MNLP)}}. The MNLP score can be expressed as follows:
\[
MNLP = -\max _{y_{1}, \ldots, y_{n}} \frac{1}{n} \sum_{i}^{n} \log \mathbb{P}\left[y_{i} | \{y_j\}\setminus y_i,\left\{\mathbf{x}_{j}\right\}\right]
\]

In our experiments, we use this normalized version of the uncertainty estimate since it has been shown to be slightly better than the classical LC \cite{shen2018}, and it is commonly applied in other works on active learning for NER.

Following \newcite{Siddhant2018DeepBA}, we implement extensions for the Transformer-based and BiLSTM-based sequence taggers applying the MC dropout technique.
\newcite{gal2016dropout} showed that applying a dropout at the prediction time allows us to consider the model as a Bayesian neural network and calculate theoretically-grounded approximations of uncertainty estimates by analyzing its multiple stochastic predictions. Like \newcite{shen2018} and \newcite{Siddhant2018DeepBA} we experiment with \textbf{\textit{variation ratio (VR)}} \cite{freeman1965elementary}: a fraction of models, which predictions differ from the majority: 
\[
VR_i = 1-\frac{\operatorname{count}\left(\operatorname{mode}\left(y^{1}_i, \ldots, y^{M}_i\right),\{y_i^m\}_m\right)}{M}
\]
\vspace{-0.3cm}
\[
VR = \frac{1}{n}\sum_i^n{VR_i},
\]
where $M$ is a number of stochastic predictions.

\newcite{Siddhant2018DeepBA} and \newcite{shen2018} refer to this method as BALD. However, the \textit{\textbf{Bayesian active learning by disagreement (BALD)}} proposed by \newcite{houlsby2011bayesian} leverage mutual information 
between outputs $y_i$ and model parameters $\theta$ trained on a dataset $\mathcal{D}$:
\[
BALD_i = H(y_i|x_i,\mathcal{D}) - E_{\theta \sim p(\theta|\mathcal{D})} \left[ H(y_i|x_i,\theta) \right]
\]
Let $p_{i}^{cm}$ be a probability of a tag $c$ for a token $i$ that is predicted by a $m$-th stochastic pass of a model with the MC dropout. Then the BALD score can be approximated according to the following expression \cite{gal2017deep}:
\[
\begin{aligned}
&BALD_i \approx \\
& -\sum_{c}^C\left(\frac{1}{M} \sum_{m}^M p_{i}^{cm}\right) \log \left(\frac{1}{M} \sum_{m}^M p_{i}^{cm}\right)\\ 
&+\frac{1}{M} \sum_{c, m}^{C,M} p_{i}^{cm} \log p_{i}^{cm}
\end{aligned}
\]
\vspace{-0.3cm}
\[
BALD \approx \frac{1}{n}\sum_{i}^{n}{BALD_i}
\]

Although BALD can be considered similar to VR, it potentially can give better uncertainty estimates than VR since it leverages the whole probability distributions produced by the model. However, this method has not been tested in the previous works on active learning for sequence tagging.

For Transformer models, we have two variants of the Monte Carlo dropout: the MC dropout on the last layer before the classification layer (MC last), and on all layers (MC all). We note that calculating uncertainty estimates in the case of MC all requires multiple stochastic passes, and in each of them, we have to perform inference of the whole Transformer model. However, if we replace the dropout only in the classifier, multiple recalculations are needed only for the classifier, while it is enough to perform the inference of the massive Transformer ``body'' only once. Therefore, in this case, the overhead of calculating the Bayesian uncertainty estimates can be less than 1\% (in the case of 10 stochastic passes for ELECTRA according to the number of parameters) compared to deterministic strategies like MNLP.

The BiLSTM-CRF model has two types of dropout: the word dropout that randomly drops entire words after the embedding layer and the locked dropout \cite{gal2016theoretically} that drops the same neurons in the embedding space of a recurrent layer for a whole sequence. Therefore, for the BiLSTM-CRF taggers, we have three options: replacing the locked dropout (MC locked), replacing the word dropout (MC word), and replacing both of them (MC all). We should note that obtaining Bayesian uncertainty estimates does not require the recalculation of word embeddings (in our case, ELMo).


\section{Experiments and Results}
\label{sec:exps}

\subsection{Experimental Setup}

We experiment with two widely-used datasets for evaluation of sequence tagging models and AL query strategies: English CoNLL-2003 \cite{conll2003} and English OntoNotes 5.0 \cite{pradhan2013towards}. The corpora statistics are presented in Appendix \ref{apdx:dataset_stats}. In the experiments, we use the ``native'' tag schemes: IOB1 for CoNLL-2003 and IOB2 for OntoNotes 5.0.

Each experiment is an \textit{emulation} of the AL cycle: selected samples are not presented to experts for annotation but are labeled automatically according to the gold standard. Each experiment is performed for each AL query strategy and is repeated five times for CoNLL-2003 and three times for OntoNotes to report the standard deviation. A random 2\% subset in tokens of the whole training set is chosen for seeding, and instances with 2\% of tokens in total are selected for annotation on each iteration. Overall, 24 AL iterations are made, so in the final iteration,  half of the training dataset (in tokens) is labeled. We do not use validation sets provided in the corpora but keep 25\% of the labeled corpus as the development set and train models from scratch on the rest. 

Details of models and the training procedure are presented in Appendix \ref{apdx:training_details}. We conduct AL experiments with the pre-selected model and training hyperparameters. Tuning hyperparameters of acquisition models during AL would drastically increase the duration of an AL iteration, which makes AL impractical. The hyperparameter optimization is reasonable for successor models. However, in preliminary AL experiments, tuning hyperparameters of successor models on the development set appeared to demonstrate an insignificant difference in the performance compared to using pre-selected hyperparameters that are fixed for all AL iterations. Therefore, since tuning hyperparameters for training a successor model on each AL iteration drastically increases the amount of computation, we fix them to the values pre-selected for experiments without AL.

The evaluation is performed using the span-based F1-score \cite{conll2003}. For query strategies based on the MC dropout, we make ten stochastic predictions.

\subsection{Results and Discussion}

\subsubsection{Training on Entire Datasets}

\begin{table}[t]

\vskip 0.15in

\begin{center}
\begin{small}
\begin{sc}

\begin{tabular}{lcc}
\toprule
 Model & CoNLL-2003 & OntoNotes \\
\midrule
CRF             &  78.2 $\pm$ NA &  79.8 $\pm$ NA \\
CNN-BiLSTM-CRF  &  88.3 $\pm$ 0.2 &  82.9 $\pm$ 0.3 \\ 
ELMo-BiLSTM-CRF &  91.2 $\pm$ 0.2 &  87.2 $\pm$ 0.2 \\ 
DistilBERT      &  89.8 $\pm$ 0.2 &  87.3 $\pm$ 0.1 \\ 
BERT            &  91.1 $\pm$ 0.2 &  88.2 $\pm$ 0.2 \\ 
ELECTRA         &  91.5 $\pm$ 0.2 &  87.6 $\pm$ 0.2 \\
\bottomrule

\end{tabular}

\end{sc}
\end{small}
\end{center}
\caption{Performance of models built on the entire training datasets without active learning}

\label{tab:fulldataset}
\end{table}

\begin{figure*}[!ht]
    \centering
    \footnotesize
    \begin{minipage}[h]{0.49\linewidth}
    \center{\includegraphics[width=\linewidth]{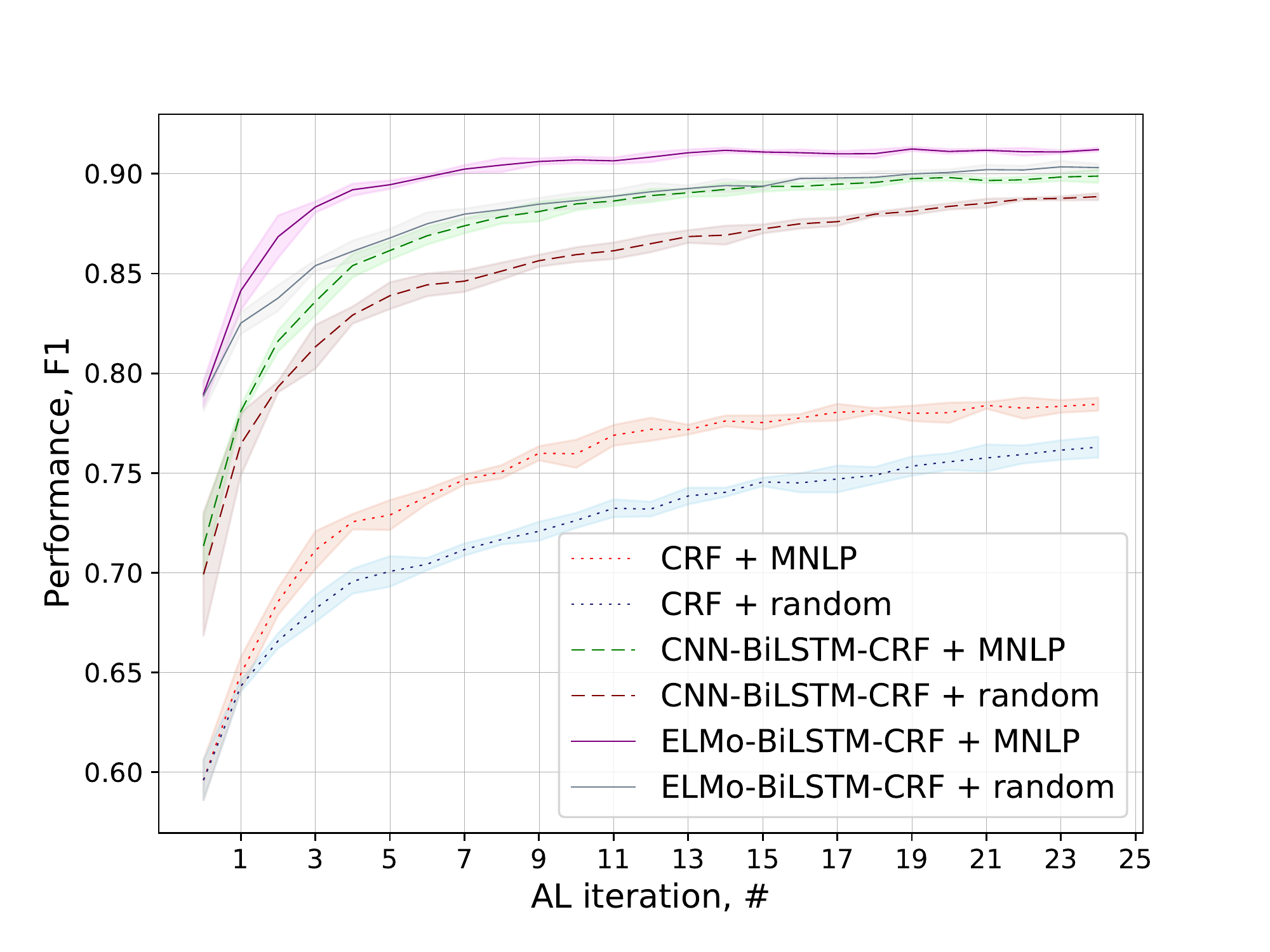} a) CoNLL-2003 \vspace{0.3cm}}
    \end{minipage}
    \hspace{0.1cm}
    \begin{minipage}[h]{0.49\linewidth}
    \center{\includegraphics[width=\linewidth]{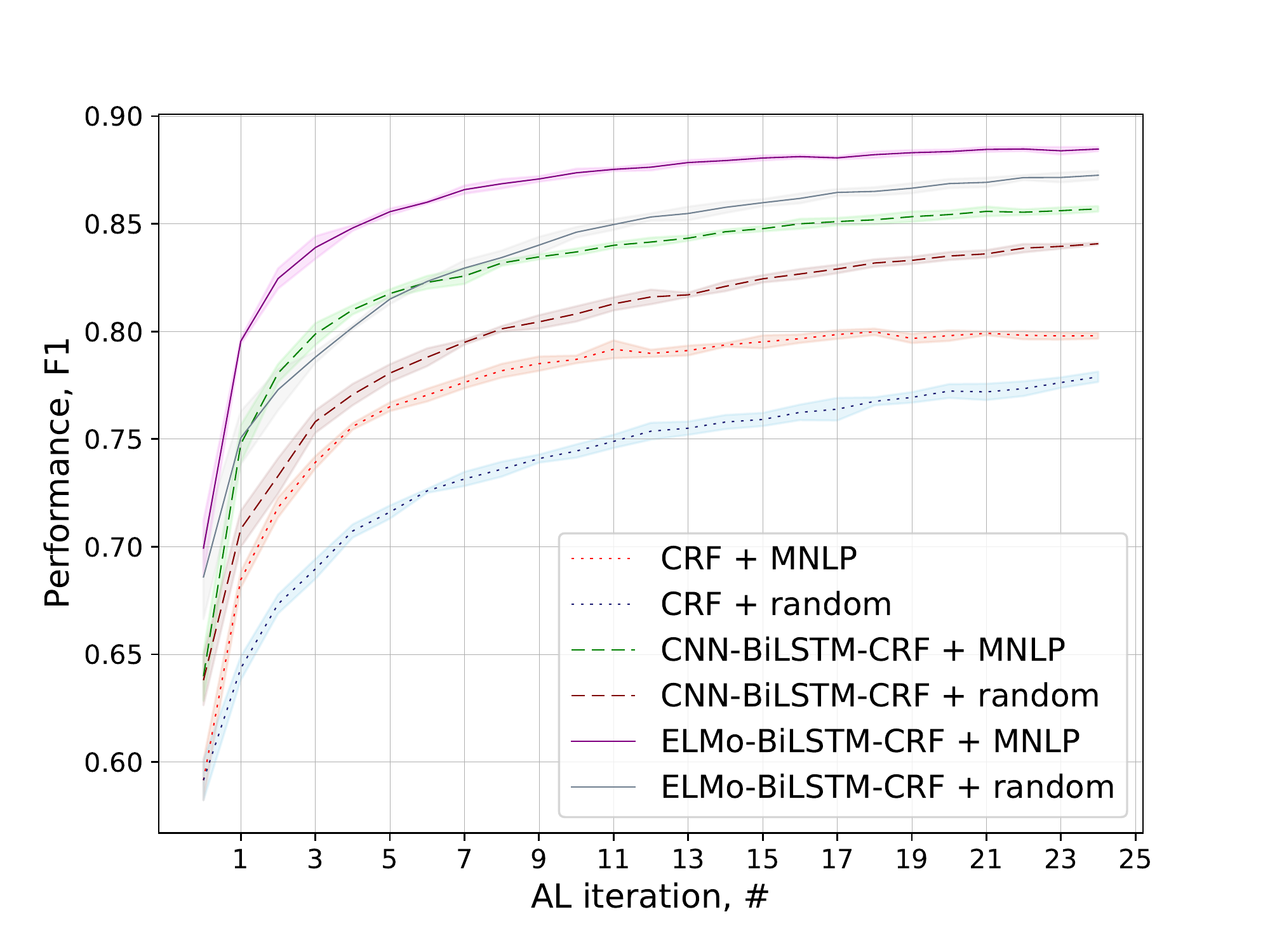} b) OntoNotes \vspace{0.3cm}}
    \end{minipage}
    The comparison of MNLP and random query strategies for CRF and BiLSTM-based models.
    \centering
    \begin{minipage}[h]{0.49\linewidth}
    \center{\includegraphics[width=\linewidth]{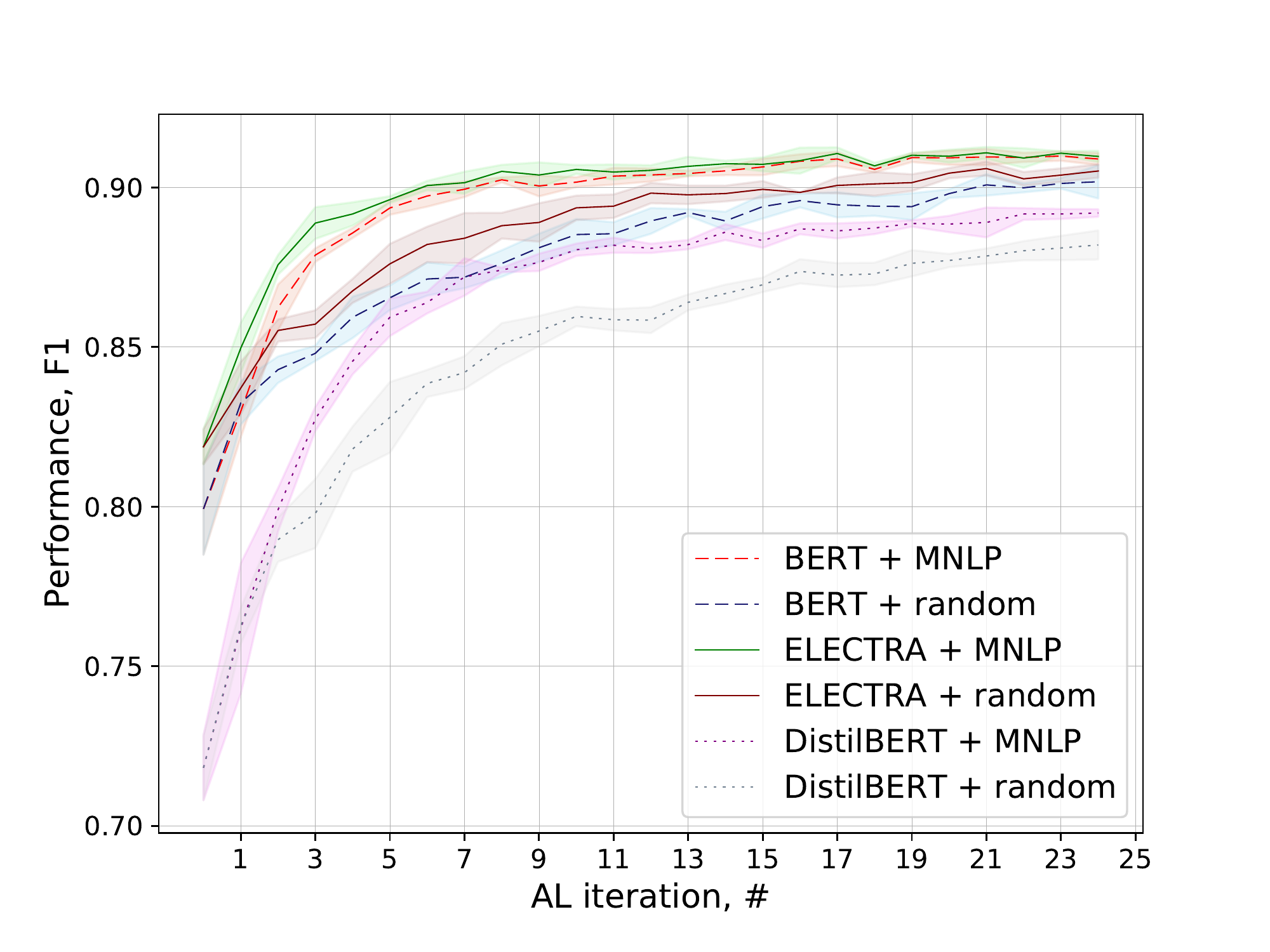} c) CoNLL-2003 \vspace{0.3cm}}
    \end{minipage}
    \hspace{0.1cm}
    \begin{minipage}[h]{0.49\linewidth}
    \center{\includegraphics[width=\linewidth]{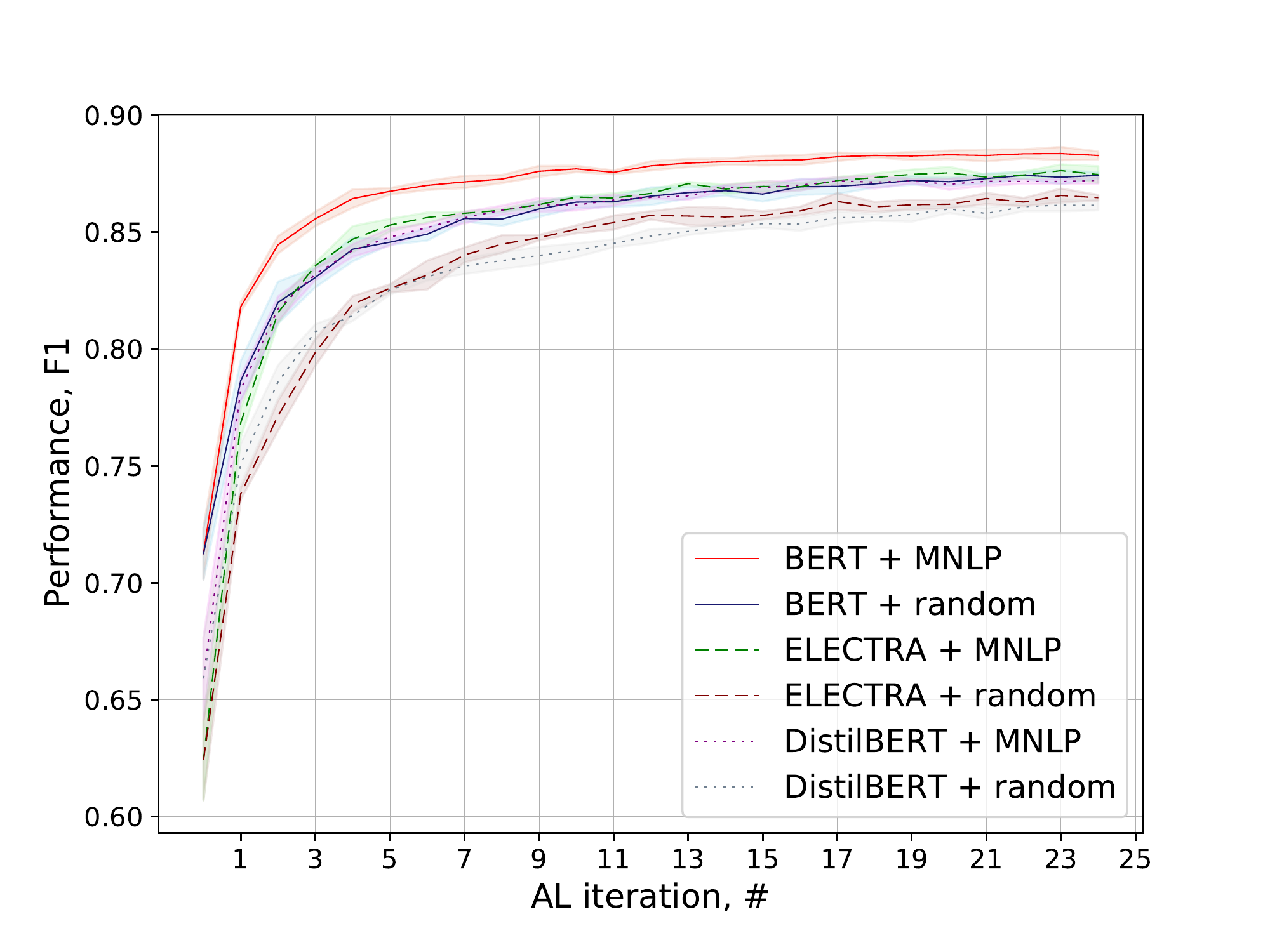} d) OntoNotes \vspace{0.3cm}}
    \end{minipage}
    The comparison of MNLP and random sampling query strategies for Transformers.
    \caption{The comparison of MNLP and random query strategies }
    \label{fig:al_mnlp}
\end{figure*}

\begin{figure*}[t]
    \footnotesize
    
    \centering
    \begin{minipage}[h]{0.49\linewidth}
    \center{\includegraphics[width=\linewidth]{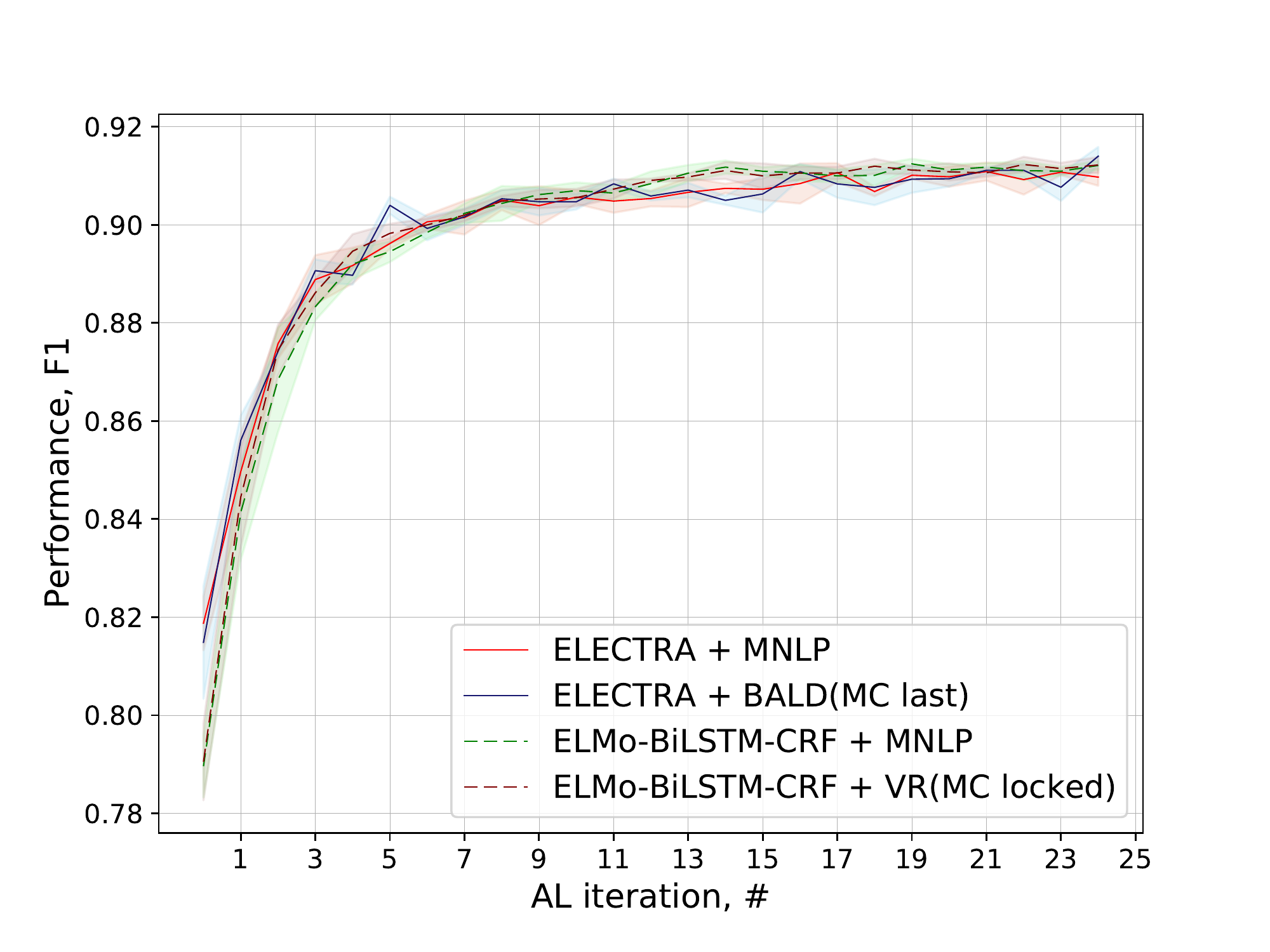} a) CoNLL}
    \end{minipage}
    \hspace{0.1cm}
    \begin{minipage}[h]{0.49\linewidth}
    \center{\includegraphics[width=\linewidth]{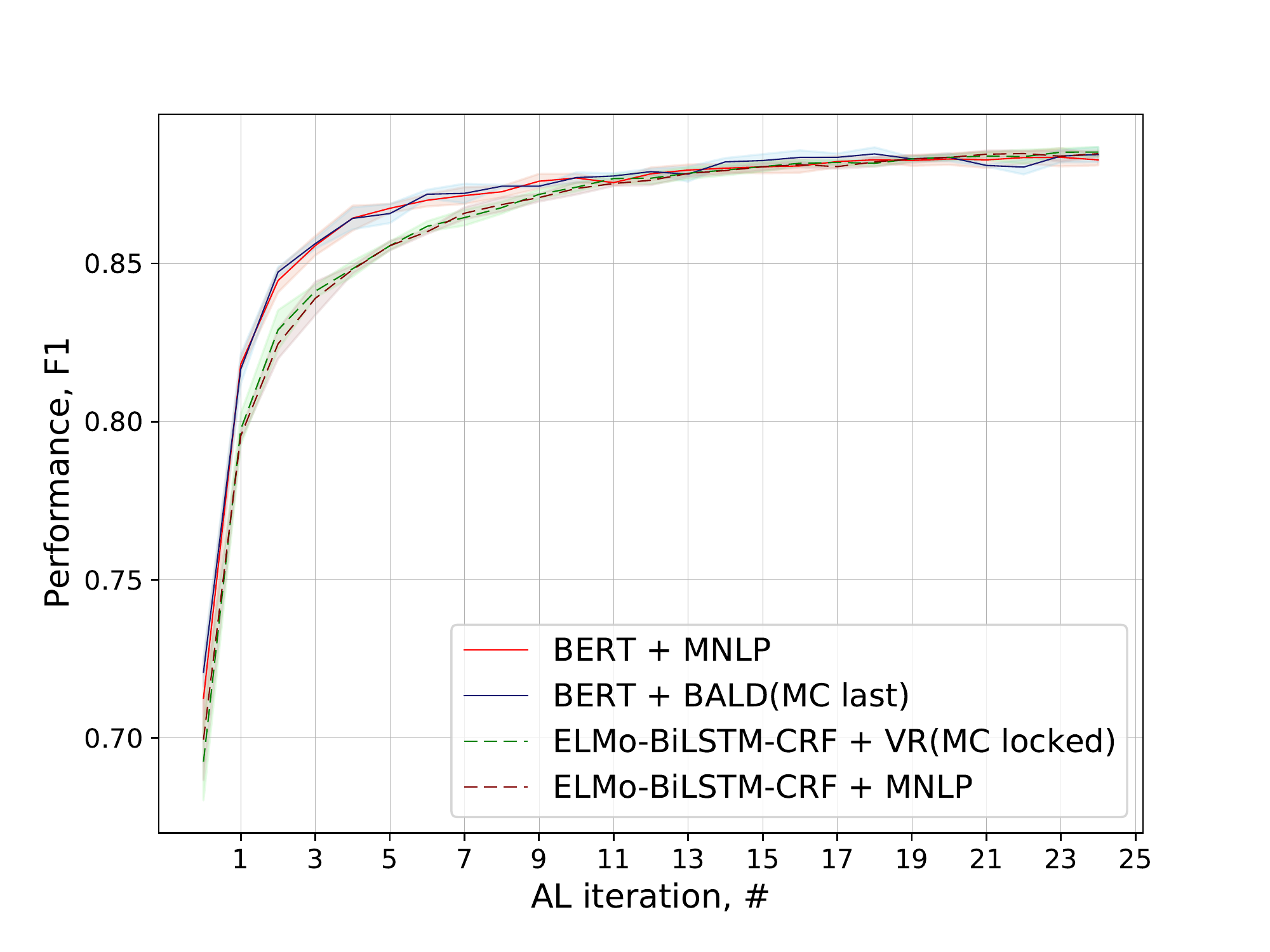} b) OntoNotes}
    \end{minipage}
    
    \caption{The comparison of the best query strategies and models overall}
    
    \label{fig:al_best}
\end{figure*}

\begin{figure*}[ht]
    \footnotesize
    
    \centering
    \begin{minipage}[h]{0.49\linewidth}
    \center{\includegraphics[width=\linewidth]{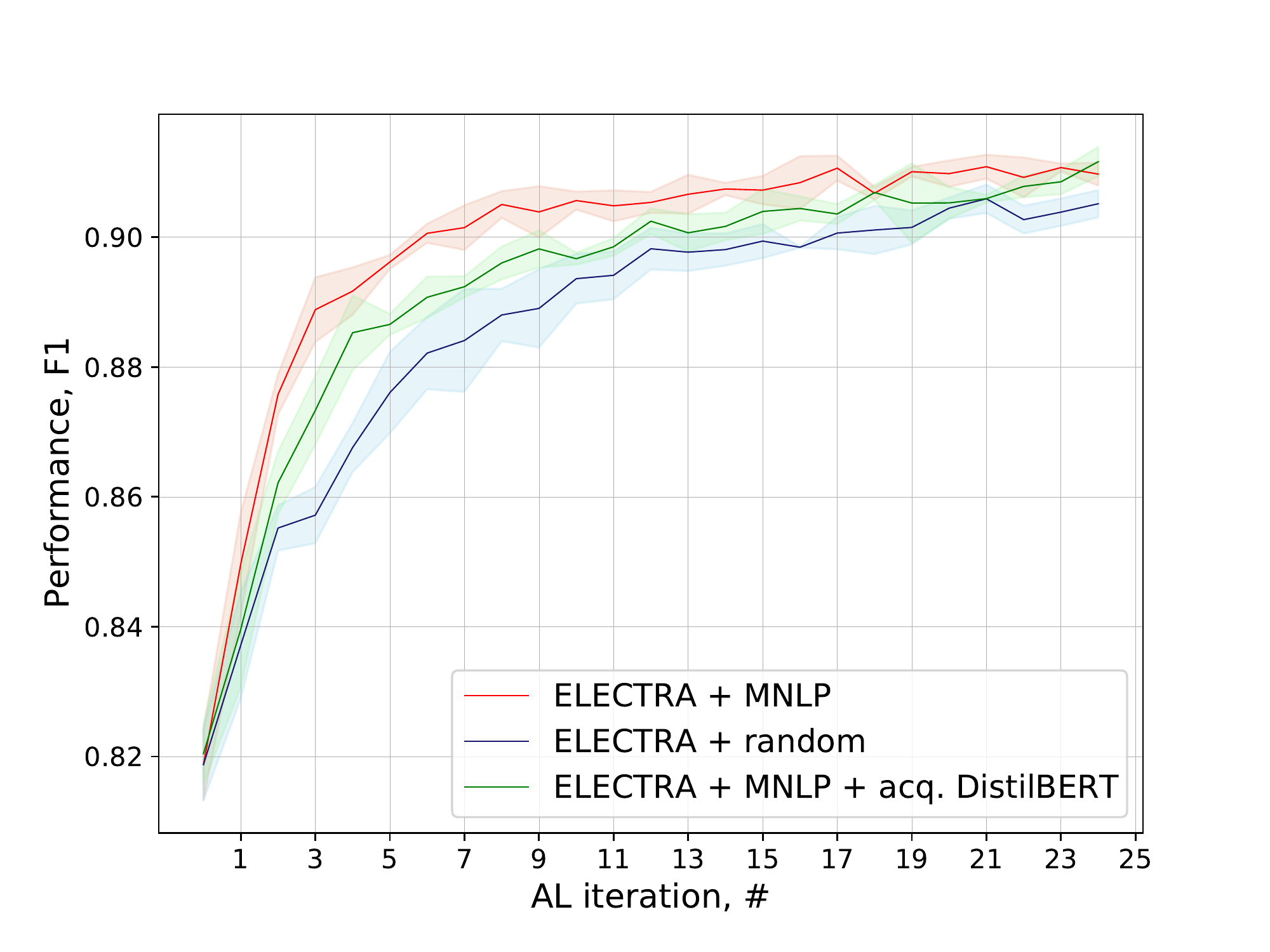} a) DistilBERT is an acquisition model}
    \end{minipage}
    \hspace{0.1cm}
    \begin{minipage}[h]{0.49\linewidth}
    \center{\includegraphics[width=\linewidth]{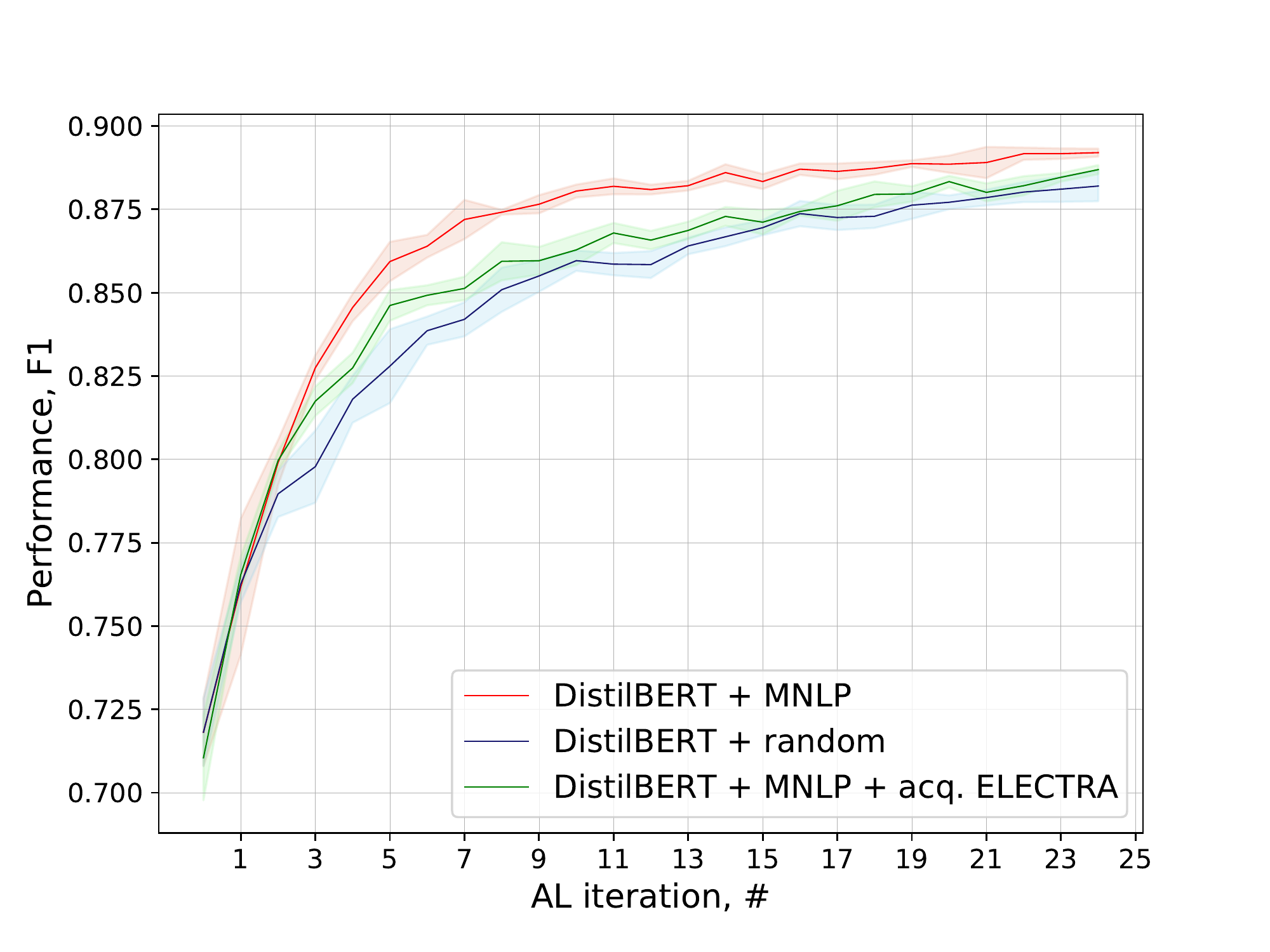} b) ELECTRA is an acquisition model \vspace{0.3cm}}
    \end{minipage}
    DistilBERT/ELECTRA as acquisition/successor models.
    
    \centering
    \begin{minipage}[h]{0.49\linewidth}
    \center{\includegraphics[width=\linewidth]{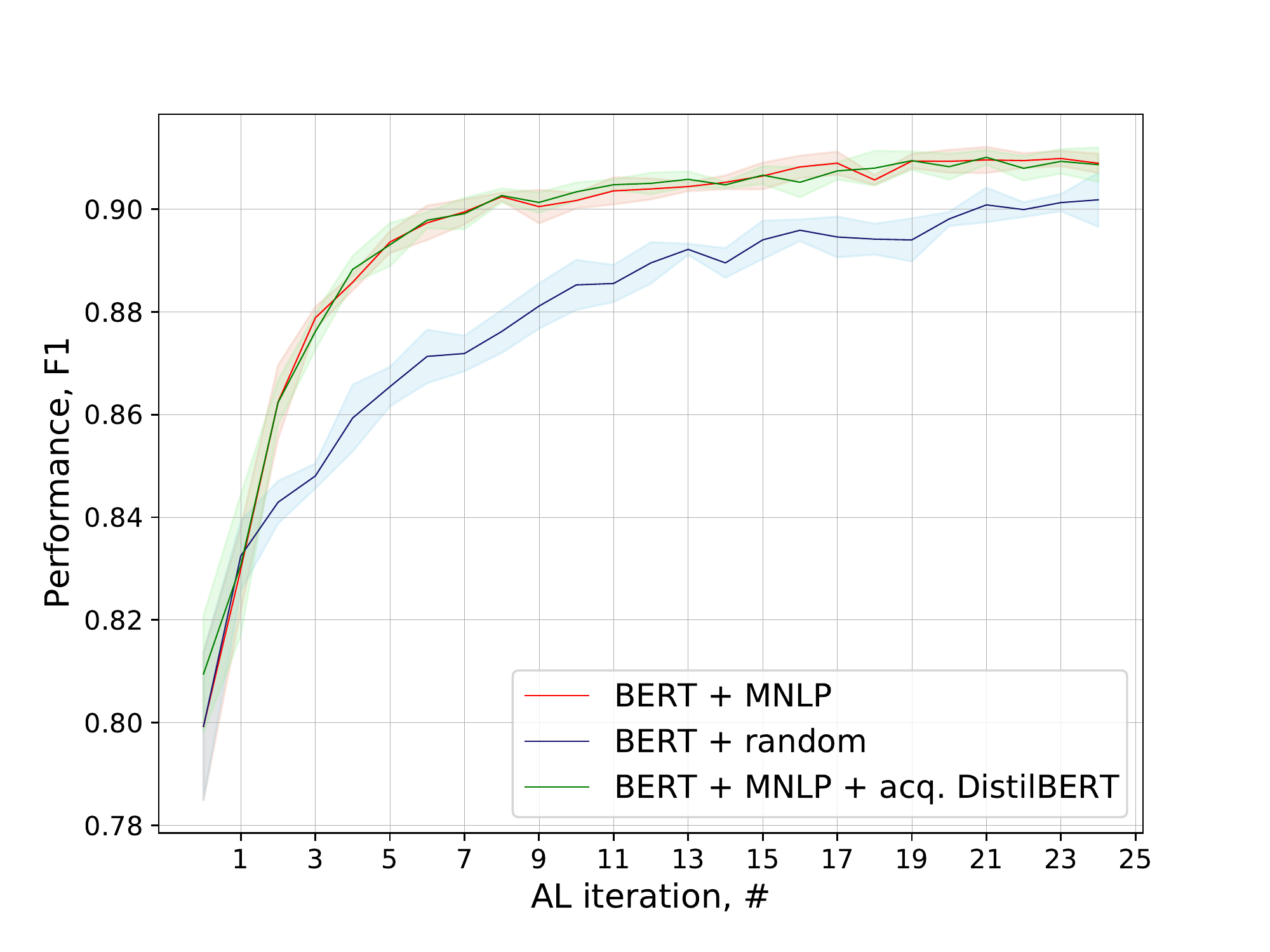} c) DistilBERT is an acquisition model \vspace{0.3cm}}
    \end{minipage}
    \hspace{0.1cm}
    \begin{minipage}[h]{0.49\linewidth}
    \center{\includegraphics[width=\linewidth]{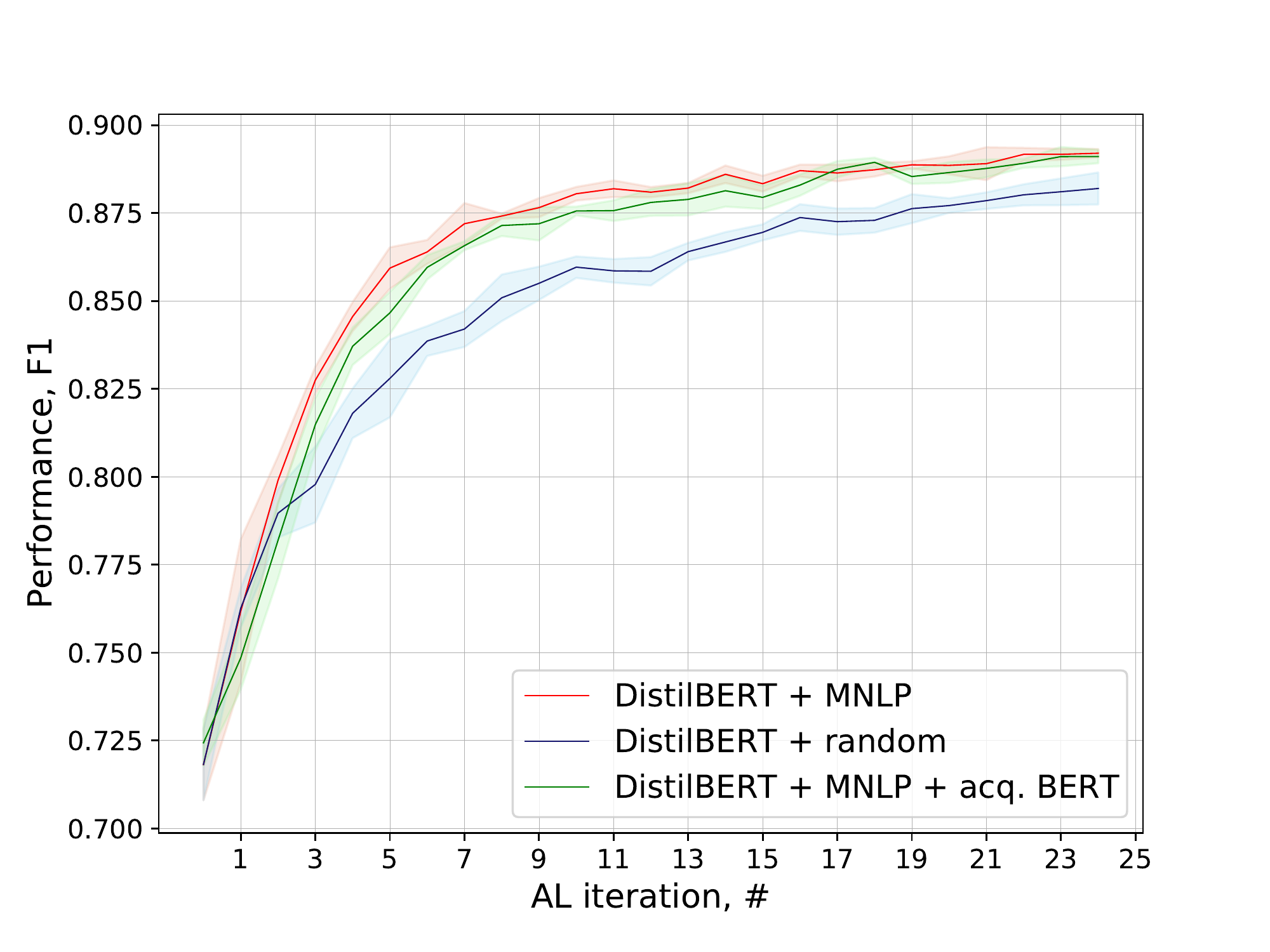} d) BERT is an acquisition model \vspace{0.3cm}}
    \end{minipage}
    DistilBERT/BERT as acquisition/successor models.
    
    \caption{AL experiments on the CoNLL-2003 dataset, in which a successor model does not match an acquisition model.}
    \label{fig:distilbert_acq}
\end{figure*}

We evaluate the performance of models trained on 75\% of the available training corpus while keeping the rest 25\% as the development set similarly to the experiments with AL.
From Table \ref{tab:fulldataset}, we can find that for both OntoNotes and CoNLL-2003, the model performance pattern is almost the same. CRF, as the baseline model, has the lowest F1 score. Sequence taggers based on deep pre-trained models achieve substantially higher results compared to the classical CNN-BiLSTM-CRF model and CRF. BERT and ELECTRA significantly outperform ELMo-BiLSTM-CRF on OntoNotes, while on CoNLL-2003, all models have comparable scores. DistilBERT is behind larger Transformers. It also has a lower performance than ELMo-BiLSTM-CRF on CoNLL-2003 but similar scores on the OntoNotes dataset. We should note that our goal was not to show the state-of-the-art performance on each dataset but to determine reasonable hyperparameters and reference scores for experiments with AL.    

\subsubsection{Active Learning}

The main results of experiments with AL are presented in Figures \ref{fig:al_mnlp}--\ref{fig:distilbert_acq}. AL shows significant improvements over the random sampling baseline for all models and on both datasets. Performance gains are bigger for simpler models like CRF or CNN-BiLSTM-CRF without deep pre-training and for the more complex OntoNotes dataset. However, we see that both ELMo-BiLSTM-CRF and Transformers benefit from a biased sampling of AL query strategies, which magnifies their ability to be trained on extremely small amount of labeled data. For example, to get 99\% of the score achieved with training on the entire CoNLL-2003 dataset, only 20\% of the annotated data is required for the ELECTRA tagger accompanied with the best AL strategy. For OntoNotes and BERT, only 16\% of the corpus is required. Random sampling requires more than 44\% and 46\% of annotated data for CoNLL-2003 and OntoNotes correspondingly.

\textbf{MNLP strategy.}
For the CoNLL-2003 corpus, the best performance in AL with the MNLP query strategy is achieved by the ELECTRA model. It shows significantly better results in the beginning compared to BERT (see Figure \ref{fig:al_mnlp}c). ELECTRA also slightly outperforms the ELMo-BiLSTM-CRF tagger in the beginning and is on par with it on the rest of the AL curve (see Figure \ref{fig:al_best}a). The CNN-BilSTM-CRF model is always better than the baseline CRF but worse than the models that take advantage of deep pre-training (see Figure \ref{fig:al_mnlp}a). DistilBERT appeared to be the weakest model in the experiments on the CoNLL-2003 dataset. With random sampling, it is on par with the baseline CNN-BiLSTM-CRF model on most iterations, but with the MNLP strategy, DistilBERT significantly falls behind CNN-BiLSTM-CRF.

Although BERT is slightly worse than the ELECTRA and ELMo-BiLSTM-CRF taggers in the experiments on the CoNLL-2003 corpus, for OntoNotes, BERT has a significant advantage over them up to 2.3 \% of the F1 score in the AL setting on early iterations. The ELMo-BiLSTM-CRF tagger falls behind the main Transformers on the OntoNotes corpus. This might be because the BiLSTM-based tagger is underfitted to the bigger corpus with only 30 training epochs. In the same vein, the baseline CNN-BiLSTM-CRF model without deep pre-training significantly falls behind DistilBERT on this corpus for MNLP and random query strategies.

\textbf{Bayesian active learning.}
Bayesian uncertainty estimates based on the MC dropout perform comparably with the deterministic MNLP strategy for Transformer-based and BiLSTM-CRF-based taggers (see Figure \ref{fig:al_best} and Tables \ref{tab:conll_bayes}, \ref{tab:ontonotes_bayes} in Appendix \ref{app:mc_dropout_options}). We consider that Bayesian uncertainty estimates do not outperform the deterministic uncertainty estimates because the performance of the latter is very close to the maximum that can be achieved with the given amount of data. 

We compare the performance of Bayesian AL strategies, when different dropout layers are replaced with the MC dropout. For ELMo-BiLSTM-CRF, we compare three options: replacing the dropout that follows word embeddings (embeddings acquired from ELMo), locked dropout in the recurrent layer, and both. Replacing the dropout that follows the embedding layer degrades the performance of AL significantly, especially for BALD. Replacing both yields the same performance as replacing only the locked dropout. We consider that the latter option is the best for AL since it requires fewer changes to the architecture. Overall, for both datasets, variation ratio has a slight advantage over the BALD strategy for the ELMo-BiLSTM-CRF model for all MC dropout options.

For Transformers, we compare two options: replacing the dropout only on the last classification layer and all dropouts in the model. When using the variation ratio strategy, replacing only the last dropout layer with the MC dropout degrades the performance compared to MNLP, while replacing all dropouts shows comparable results with it. However, for the BALD strategy, we see the inverse situation: replacing the last dropout layer leads to the significantly better performance than replacing all layers. This pattern can be noted for both ELECTRA and DistilBERT on the CoNLL-2003 corpus and for both BERT and DistilBERT on the OntoNontes corpus. Therefore, for Transformers, BALD with the MC dropout on the last layer is the best Bayesian query strategy since it provides both good quality and low computational overhead.

\begin{table}[t]
\footnotesize

\caption{Duration of active learning iteration phases (seconds). The presented values correspond to the 5-th iteration of AL on the CoNLL-2003 dataset with the MNLP query strategy. The duration is averaged over three runs.}
\label{tab:duration}

\begin{center}
\begin{sc}

\resizebox{\columnwidth}{!}{%
\begin{tabular}{lccc}
\toprule
{Model} &    \multirowcell{Acq. model \\training} &  \multirowcell{Querying \\ inst.} & Total \\
\midrule
CNN-BiLSTM-CRF    & 1052 & 68  & 1120 \\
ELMo-BiLSTM-CRF   & 339  & 100 & 439  \\
ELECTRA           & 100  & 50  & 150  \\
DistilBERT        & 61   & 27  & 88   \\
\bottomrule
\end{tabular}
}

\end{sc}
\end{center}
\end{table}

\textbf{Duration of AL iterations.}
The valuable benefit of deep pre-trained models is also their high inference and training speed, which helps to significantly reduce the duration of an AL iteration and makes it feasible to implement text annotation tools empowered with interactive deep active learning. We measure the duration of the acquisition model training phase and the duration of the instance querying phase, which includes model inference on the whole unlabeled dataset (see Table \ref{tab:duration}). The experiments were conducted using the Nvidia V100 GPU and Xeon E5-2698 CPU.

According to obtained results, the large ELECTRA model can be trained more than ten times faster than the basic CNN-BiLSTM-CRF model because of a hardware-optimized architecture, a smaller number of the necessary training epochs, and the absence of the validation on the development set. ELECTRA also helps to reduce the duration of the query selection by more than 26\%. The lightweight DistilBERT model is expectedly even faster. We also should note that using BiLSTM-CRF with the pre-trained contextualized word representation model ELMo also helps to make the duration of AL iterations shorter compared to CNN-BiLSTM-CRF due to no need for training of the CNN word representation subnetwork (ELMo usually is not fine-tuned) and caching of the forward passes of ELMo during training across different epochs.

\textbf{Mismatch between a successor model and an acquisition model.} 
Since AL creates substantial computational overhead caused by training and inference of an acquisition model, for practical usage, it is reasonable to keep an acquisition model as lightweight as possible, while retaining the successor model complex and full-fledged. However, \newcite{lowell2019practical} demonstrate that the mismatch between an acquisition model and a successor model can diminish the benefits of applying the AL. They show that using AL can even harm the performance of the successor model compared to random sampling. We investigate the situation when an acquisition model does not match a successor model. Figure \ref{fig:distilbert_acq} shows the results of BERT, DistilBERT, and ELECTRA in this setting.  

From Figures \ref{fig:distilbert_acq}a and \ref{fig:distilbert_acq}b, we can find that, when DistilBERT is used as an acquisition model for ELECTRA and vice versa, improvements achieved due to AL over the random sampling baseline are substantially lower compared to the ``native'' acquisition model. The same negative effect can be seen due to a mismatch between ELECTRA and BERT models in Figure \ref{fig:bert_electra_acq} in Appendix \ref{app:mismatch}. These results support the findings presented in \cite{lowell2019practical}. However, in our case, AL still gives notable improvements over the random sampling baseline. From Figure \ref{fig:distilbert_acq}d, we can see that using BERT as an acquisition model for DistilBERT results only in a slight reduction of performance compared to using the ``native'' acquisition model. Moreover, in a reversed experiment, when DistilBERT is used as an acquisition model (see Figure \ref{fig:distilbert_acq}c), there is no performance drop in AL at all. Such results for BERT-DistilBERT can be explained by the relationship between the distilled model and its ancestor resulting in similar uncertainty estimates for unlabeled instances in the annotation pool. This finding reveals the possibility of replacing a big acquisition model such as BERT with a distilled version that is faster and requires much less amount of memory. This can help to alleviate practical obstacles of deploying AL in real-world scenarios.


\section{Conclusion}
\label{sec:conclusion}

In this work, we investigated the combination of AL with sequence taggers that take advantage of deep pre-trained models. In the AL setting, these sequence taggers substantially outperform the models that do not use deep pre-training. We show that AL and transfer learning is a very powerful combination that can help to produce remarkably performing models with just a small fraction of the annotated data. For the CoNLL-2003 corpus, the combination of the best performing pre-trained model and AL strategy achieves 99\% of the score that can be obtained with training on the full corpus, while using only 20\% of the annotated data. For the OntoNotes corpus, one needs just 16\%.

We performed a large empirical study of AL query strategies based on the Monte Carlo dropout in conjunction with deep pre-trained models and are the first to apply Bayesian active learning by disagreement to sequence tagging tasks. Bayesian active learning by disagreement achieves better results than the variation ratio for Transformers. However, we find that the variation ratio is slightly better for the ELMo-BiLSTM-CRF model. It is reasonable to use both MC dropout-based query strategies when only the last dropout layer works in a stochastic mode during the inference. This makes this type of query strategies suitable for practical usage due to little computational overhead. Finally, we demonstrate that it is possible to reduce the computational overhead of AL with deep pre-trained models by using a smaller distilled version of a Transformer model for acquiring instances. 

In the future work, we are seeking to extend the empirical investigation of deep pre-trained models in active learning to query strategies that aim at better handling the batch selection of instances for annotation.


\section*{Acknowledgments}
We thank the reviewers for their valuable feedback. This work was done in the framework of the joint MTS-Skoltech lab. The development of the experimental setup for the study of active learning methods and its application to  sequence tagging  tasks (Section 4) was supported by the Russian Science Foundation grant 20-71-10135. The Zhores supercomputer~\cite{Zhores} was used for computations. We are very grateful to Sergey Ustyantsev for help with the implementation and the execution of the experiments.

\bibliography{references}
\bibliographystyle{acl_natbib}

\clearpage
\newpage
\appendix

\section{Dataset Characteristics}
\label{apdx:dataset_stats}

Tables \ref{tab:ontonotes_dataset} and \ref{tab:dataset_conll2003} present the characteristics of the datasets used in experiments.

\begin{table}[H]
\footnotesize

\caption{Characteristics of the OntoNotes 5.0 corpus (without the PT section)}
\label{tab:ontonotes_dataset}

\begin{center}
\begin{sc}
\begin{tabular}{lccr}
\toprule
{} &    ENG.TRAIN &    ENG.TEST \\
\midrule
\# OF TOKENS    &  1,088,503 &  152,728 \\
\# OF SENTENCES &   59,924 &   8,262 \\
\midrule
Entity types: \\
\midrule
PERSON      &  15,429 &  1,988 \\
GPE         &  15,405 &  2,240 \\
ORG         &  12,820 &  1,795 \\
DATE        &  10,922 &  1,602 \\
CARDINAL    &   7,367 &    935 \\
NORP        &   6,870 &    841 \\
MONEY       &   2,434 &    314 \\
PERCENT     &   1,763 &    349 \\
ORDINAL     &   1,640 &    195 \\
LOC         &   1,514 &    179 \\
TIME        &   1,233 &    212 \\
WORK\_OF\_ART &     974 &    166 \\
FAC         &     860 &    135 \\
EVENT       &     748 &     63 \\
QUANTITY    &     657 &    105 \\
PRODUCT     &     606 &     76 \\
LANGUAGE    &     304 &     22 \\
LAW         &     282 &     40 \\
\midrule
Total entities: &   81,828 &   11,257 \\
\bottomrule
\end{tabular}
\end{sc}
\end{center}
\end{table}

\begin{table}[h]
\footnotesize

\caption{Characteristics of the CoNLL-2003 corpus}
\label{tab:dataset_conll2003}

\begin{center}

\begin{sc}
\begin{tabular}{lccr}
\toprule
{} &   ENG.TRAIN &   ENG.TESTB \\
\midrule
\# OF TOKENS    &  203,621 &  46,435 \\
\# OF SENTENCES &   14,041 &   3,453 \\
\midrule
Entity types: \\
\midrule
LOC            &    7,140 &   1,668 \\
PER            &   6,600 &   1,617 \\
ORG            &   6,321 &   1,661 \\
MISC           &   3,438 &    702 \\
\midrule
Total entities: &   23,499 &   5,648 \\
\bottomrule
\end{tabular}
\end{sc}

\end{center}

\end{table}


\section{Features Used by the CRF Model}
\label{appendixA}

 \begin{enumerate}[itemsep=1mm, parsep=0pt]
     \item A lowercased word form.
     \item Trigram and bigram suffixes of words.
     \item Capitalization features.
     \item An indicator that shows whether a word is a digit.
     \item A part-of-speech tag of a word with specific info (plurality, verb tense, etc.)
     \item A generalized part-of-speech.
     \item An indicator whether a word is at the beginning or ending of a sentence.
     \item The aforementioned characteristics for the next word and previous word except suffixes.
 \end{enumerate}


\section{Model and Training Details}
\label{apdx:training_details}

\subsection{CRF}

We set CRF L1 and L2 regularization terms equal to 0.1, and limit the number of iterations by 100. 

\subsection{BiLSTM-CRF Taggers}

We implement the BiLSTM-CRF sequence tagger on the basis of the \texttt{Flair} package\footnote{\url{https://github.com/flairNLP/flair}} \cite{akbik2018coling}. We use the same parameters for both types of BiLSTM-CRF models. The recurrent network has one layer with 128 neurons. During training, we anneal the learning rate by half, when the performance of the model stops improving on the development set for 3 epochs. After annealing, we restore the model from the epoch with the best validation score. The starting learning rate is 0.1. The maximal number of epochs is 30, and the batch size is 32. For optimization, we use the standard SGD algorithm.

\subsection{Transformer-based Taggers}

The implementation of Trasnformer-based taggers is based on the \texttt{Hugging Face Transformers} \cite{Wolf2019HuggingFacesTS}\footnote{\url{https://huggingface.co/transformers/}} library. We use the following pre-trained versions of BERT, ELECTRA, and DistilBERT accordingly: \textit{`bert-base-cased'}, \textit{`google/electra-base-discriminator'}, and \textit{`distilbert-base-cased'}. The corrected version of Adam (AdamW from the \texttt{Transformers} library) is used for optimization with the base learning rate of 5e-5. The linear decay of the learning rate is applied following the \cite{Devlin2019BERTPO}. The number of epochs is 4 and the batch size is 16. As in \cite{shen2018}, we see that it is critical to adjust the batch size on early AL iterations, when only small amount of labeled data is available. We reduce the batch size to keep the number of iterations per epoch over 50, but limit the minimal batch size to 4. 

\clearpage


\newcommand{\specialcell}[2][c]{%
  \begin{tabular}[#1]{@{}c@{}}#2\end{tabular}}

\onecolumn
\section{Comparison of Various MC Dropout Options}
\label{app:mc_dropout_options}

\begin{table*}[h!]

\footnotesize

\caption{Results of AL with various MC dropout options on the CoNLL-2003 dataset}
\label{tab:conll_bayes}
\centering

\begin{tabular}{llcccccc}
\toprule
                &        &          1  &          5  &          10 &          15 &          20 &          24 \\
\midrule
ELMo-BiLSTM-CRF & MNLP &  84.1 $\pm$ 1.0 &  89.5 $\pm$ 0.2 &  90.7 $\pm$ 0.2 &  91.1 $\pm$ 0.1 &  91.1 $\pm$ 0.1 &  91.2 $\pm$ 0.1 \\
        & Random &  82.5 $\pm$ 0.5 &  86.8 $\pm$ 0.4 &  88.7 $\pm$ 0.4 &  89.4 $\pm$ 0.2 &  90.1 $\pm$ 0.2 &  90.3 $\pm$ 0.2 \\
        & VR(MC word) &  83.3 $\pm$ 1.0 &  89.4 $\pm$ 0.1 &  90.5 $\pm$ 0.1 &  90.8 $\pm$ 0.3 &  91.0 $\pm$ 0.2 &  91.1 $\pm$ 0.1 \\
        & VR(MC all) &  84.7 $\pm$ 0.7 &  89.7 $\pm$ 0.2 &  90.7 $\pm$ 0.1 &  90.9 $\pm$ 0.1 &  90.9 $\pm$ 0.1 &  91.2 $\pm$ 0.2 \\
        & VR(MC locked) &  84.4 $\pm$ 1.0 &  89.8 $\pm$ 0.2 &  90.6 $\pm$ 0.2 &  91.0 $\pm$ 0.3 &  91.1 $\pm$ 0.2 &  91.2 $\pm$ 0.2 \\
        & BALD(MC word) &  83.5 $\pm$ 1.0 &  88.8 $\pm$ 0.5 &  90.3 $\pm$ 0.2 &  90.5 $\pm$ 0.2 &  90.9 $\pm$ 0.2 &  91.1 $\pm$ 0.2 \\
        & BALD(MC locked) &  84.3 $\pm$ 0.4 &  89.5 $\pm$ 0.4 &  90.5 $\pm$ 0.0 &  90.8 $\pm$ 0.1 &  91.1 $\pm$ 0.1 &  91.3 $\pm$ 0.1 \\
        & BALD(MC all) &  84.2 $\pm$ 0.7 &  89.6 $\pm$ 0.2 &  90.5 $\pm$ 0.2 &  90.7 $\pm$ 0.1 &  91.1 $\pm$ 0.1 &  91.1 $\pm$ 0.1 \\
\hline
DistilBERT & MNLP &  76.2 $\pm$ 2.0 &  85.9 $\pm$ 0.6 &  88.0 $\pm$ 0.2 &  88.3 $\pm$ 0.2 &  88.9 $\pm$ 0.3 &  89.2 $\pm$ 0.1 \\
        & Random &  76.3 $\pm$ 0.6 &  82.8 $\pm$ 1.1 &  86.0 $\pm$ 0.3 &  87.0 $\pm$ 0.2 &  87.7 $\pm$ 0.2 &  88.2 $\pm$ 0.5 \\
        & VR (MC last) &  78.0 $\pm$ 0.2 &  85.3 $\pm$ 0.2 &  87.7 $\pm$ 0.5 &  88.1 $\pm$ 0.4 &  88.6 $\pm$ 0.5 &  89.0 $\pm$ 0.3 \\
        & VR (MC all) &  78.3 $\pm$ 1.4 &  86.3 $\pm$ 0.6 &  88.3 $\pm$ 0.1 &  88.8 $\pm$ 0.3 &  88.8 $\pm$ 0.3 &  89.0 $\pm$ 0.2 \\
        & BALD (MC last) &  78.0 $\pm$ 0.9 &  86.0 $\pm$ 0.2 &  87.7 $\pm$ 0.3 &  88.3 $\pm$ 0.3 &  89.1 $\pm$ 0.0 &  89.1 $\pm$ 0.1 \\
        & BALD (MC all) &  77.3 $\pm$ 1.0 &  85.6 $\pm$ 0.4 &  87.9 $\pm$ 0.1 &  88.6 $\pm$ 0.3 &  89.0 $\pm$ 0.1 &  89.0 $\pm$ 0.1 \\
\hline
ELECTRA & MNLP &  85.0 $\pm$ 0.8 &  89.6 $\pm$ 0.1 &  90.6 $\pm$ 0.1 &  90.7 $\pm$ 0.2 &  91.0 $\pm$ 0.2 &  91.0 $\pm$ 0.2 \\
        & Random &  83.7 $\pm$ 0.8 &  87.6 $\pm$ 0.6 &  89.4 $\pm$ 0.4 &  89.9 $\pm$ 0.3 &  90.4 $\pm$ 0.2 &  90.5 $\pm$ 0.2 \\
        & VR (MC last) &  85.9 $\pm$ 0.7 &  88.7 $\pm$ 0.2 &  90.3 $\pm$ 0.2 &  90.6 $\pm$ 0.5 &  91.0 $\pm$ 0.3 &  91.0 $\pm$ 0.1 \\
        & VR (MC all) &  84.3 $\pm$ 1.0 &  89.8 $\pm$ 0.2 &  90.4 $\pm$ 0.3 &  90.9 $\pm$ 0.2 &  91.0 $\pm$ 0.1 &  91.2 $\pm$ 0.3 \\
        & BALD (MC last) &  85.6 $\pm$ 0.5 &  90.4 $\pm$ 0.2 &  90.5 $\pm$ 0.2 &  90.6 $\pm$ 0.4 &  90.9 $\pm$ 0.2 &  91.4 $\pm$ 0.2 \\
        & BALD (MC all) &  84.0 $\pm$ 0.9 &  89.5 $\pm$ 0.4 &  90.6 $\pm$ 0.2 &  90.7 $\pm$ 0.3 &  91.1 $\pm$ 0.2 &  90.8 $\pm$ 0.3 \\
\bottomrule
\end{tabular}
\end{table*}

\begin{table*}[h!]
\footnotesize

\caption{Results of AL with various MC dropout options on the OntoNotes 5.0 dataset}
\label{tab:ontonotes_bayes}
\centering

\begin{tabular}{llcccccc}
\toprule
Model                 & Query strat.       &          1  &          5  &          10 &          15 &          20 &          24 \\
\midrule
ELMo-BiLSTM-CRF & MNLP &  79.5 $\pm$ 0.1 &  85.6 $\pm$ 0.2 &  87.4 $\pm$ 0.2 &  88.1 $\pm$ 0.1 &  88.4 $\pm$ 0.1 &  88.5 $\pm$ 0.1 \\
     & Random &  75.1 $\pm$ 1.2 &  81.5 $\pm$ 0.2 &  84.6 $\pm$ 0.2 &  86.0 $\pm$ 0.2 &  86.9 $\pm$ 0.2 &  87.3 $\pm$ 0.2 \\
     & VR(MC word) &  79.4 $\pm$ 0.6 &  85.1 $\pm$ 0.3 &  87.0 $\pm$ 0.1 &  87.7 $\pm$ 0.1 &  88.0 $\pm$ 0.2 &  88.2 $\pm$ 0.2 \\
     & VR(MC all) &  79.6 $\pm$ 1.0 &  85.6 $\pm$ 0.1 &  87.4 $\pm$ 0.2 &  88.0 $\pm$ 0.3 &  88.3 $\pm$ 0.1 &  88.4 $\pm$ 0.1 \\
     & VR(MC locked) &  79.8 $\pm$ 0.5 &  85.6 $\pm$ 0.1 &  87.4 $\pm$ 0.1 &  88.1 $\pm$ 0.2 &  88.4 $\pm$ 0.1 &  88.5 $\pm$ 0.2 \\
     & BALD(MC word) &  78.6 $\pm$ 0.8 &  84.4 $\pm$ 0.3 &  86.3 $\pm$ 0.4 &  87.4 $\pm$ 0.3 &  87.9 $\pm$ 0.2 &  88.2 $\pm$ 0.0 \\
     & BALD(MC locked) &  78.8 $\pm$ 1.5 &  85.1 $\pm$ 0.5 &  87.0 $\pm$ 0.3 &  87.9 $\pm$ 0.2 &  88.2 $\pm$ 0.2 &  88.4 $\pm$ 0.1 \\
     & BALD(MC all) &  78.9 $\pm$ 1.3 &  85.0 $\pm$ 0.5 &  87.0 $\pm$ 0.6 &  87.9 $\pm$ 0.3 &  88.1 $\pm$ 0.2 &  88.3 $\pm$ 0.2 \\
\hline
DistilBERT & MNLP &  78.3 $\pm$ 0.5 &  84.8 $\pm$ 0.4 &  86.2 $\pm$ 0.2 &  86.9 $\pm$ 0.2 &  87.0 $\pm$ 0.2 &  87.2 $\pm$ 0.1 \\
     & Random &  75.1 $\pm$ 0.9 &  82.5 $\pm$ 0.2 &  84.2 $\pm$ 0.3 &  85.4 $\pm$ 0.2 &  86.0 $\pm$ 0.2 &  86.1 $\pm$ 0.2 \\
     & VR (MC last) &  77.8 $\pm$ 0.3 &  84.4 $\pm$ 0.3 &  85.8 $\pm$ 0.0 &  86.6 $\pm$ 0.3 &  87.0 $\pm$ 0.3 &  87.0 $\pm$ 0.3 \\
     & VR (MC all) &  78.5 $\pm$ 0.2 &  84.6 $\pm$ 0.3 &  86.1 $\pm$ 0.0 &  86.7 $\pm$ 0.2 &  87.2 $\pm$ 0.1 &  87.2 $\pm$ 0.1 \\
     & BALD (MC last) &  78.3 $\pm$ 0.3 &  84.9 $\pm$ 0.1 &  86.3 $\pm$ 0.0 &  86.9 $\pm$ 0.0 &  87.1 $\pm$ 0.2 &  87.0 $\pm$ 0.1 \\
     & BALD (MC all) &  78.3 $\pm$ 0.1 &  84.8 $\pm$ 0.1 &  86.4 $\pm$ 0.2 &  87.2 $\pm$ 0.1 &  87.1 $\pm$ 0.1 &  87.2 $\pm$ 0.2 \\
\hline
BERT & MNLP &  81.8 $\pm$ 0.2 &  86.7 $\pm$ 0.1 &  87.7 $\pm$ 0.1 &  88.1 $\pm$ 0.2 &  88.3 $\pm$ 0.2 &  88.3 $\pm$ 0.2 \\
     & Random &  78.7 $\pm$ 0.8 &  84.6 $\pm$ 0.1 &  86.3 $\pm$ 0.3 &  86.6 $\pm$ 0.3 &  87.2 $\pm$ 0.2 &  87.4 $\pm$ 0.1 \\
     & VR (MC last) &  81.6 $\pm$ 0.8 &  86.4 $\pm$ 0.3 &  87.4 $\pm$ 0.2 &  87.7 $\pm$ 0.2 &  88.0 $\pm$ 0.4 &  88.1 $\pm$ 0.1 \\
     & VR (MC all) &  82.2 $\pm$ 0.5 &  86.8 $\pm$ 0.3 &  87.7 $\pm$ 0.2 &  88.0 $\pm$ 0.2 &  88.5 $\pm$ 0.2 &  88.5 $\pm$ 0.2 \\
     & BALD (MC last) &  81.7 $\pm$ 0.5 &  86.6 $\pm$ 0.3 &  87.7 $\pm$ 0.2 &  88.3 $\pm$ 0.2 &  88.4 $\pm$ 0.1 &  88.4 $\pm$ 0.2 \\
     & BALD (MC all) &  82.3 $\pm$ 0.7 &  86.7 $\pm$ 0.0 &  87.8 $\pm$ 0.2 &  88.3 $\pm$ 0.2 &  88.4 $\pm$ 0.1 &  88.6 $\pm$ 0.3 \\
\bottomrule
\end{tabular}

\end{table*}


\clearpage
\section{Experiments with a Mismatch between a Successor Model and an Acquisition Model}
\label{app:mismatch}

\begin{figure*}[ht]
    \footnotesize
    \centering
    \begin{minipage}[h]{0.49\linewidth}
    \center{\includegraphics[width=\linewidth]{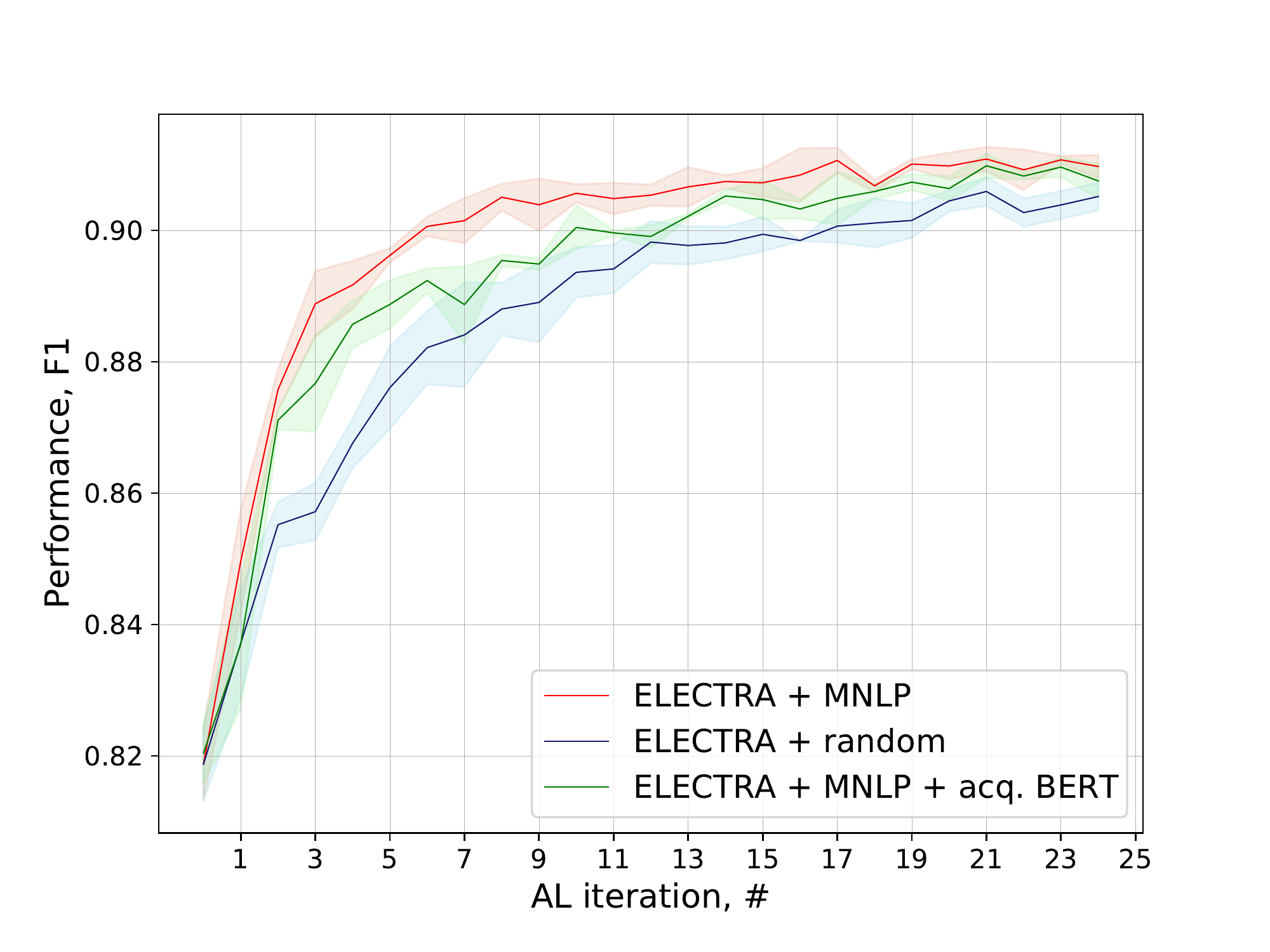} a) BERT is an acquisition model}
    \end{minipage}
    \hspace{0.1cm}
    \begin{minipage}[h]{0.49\linewidth}
    \center{\includegraphics[width=\linewidth]{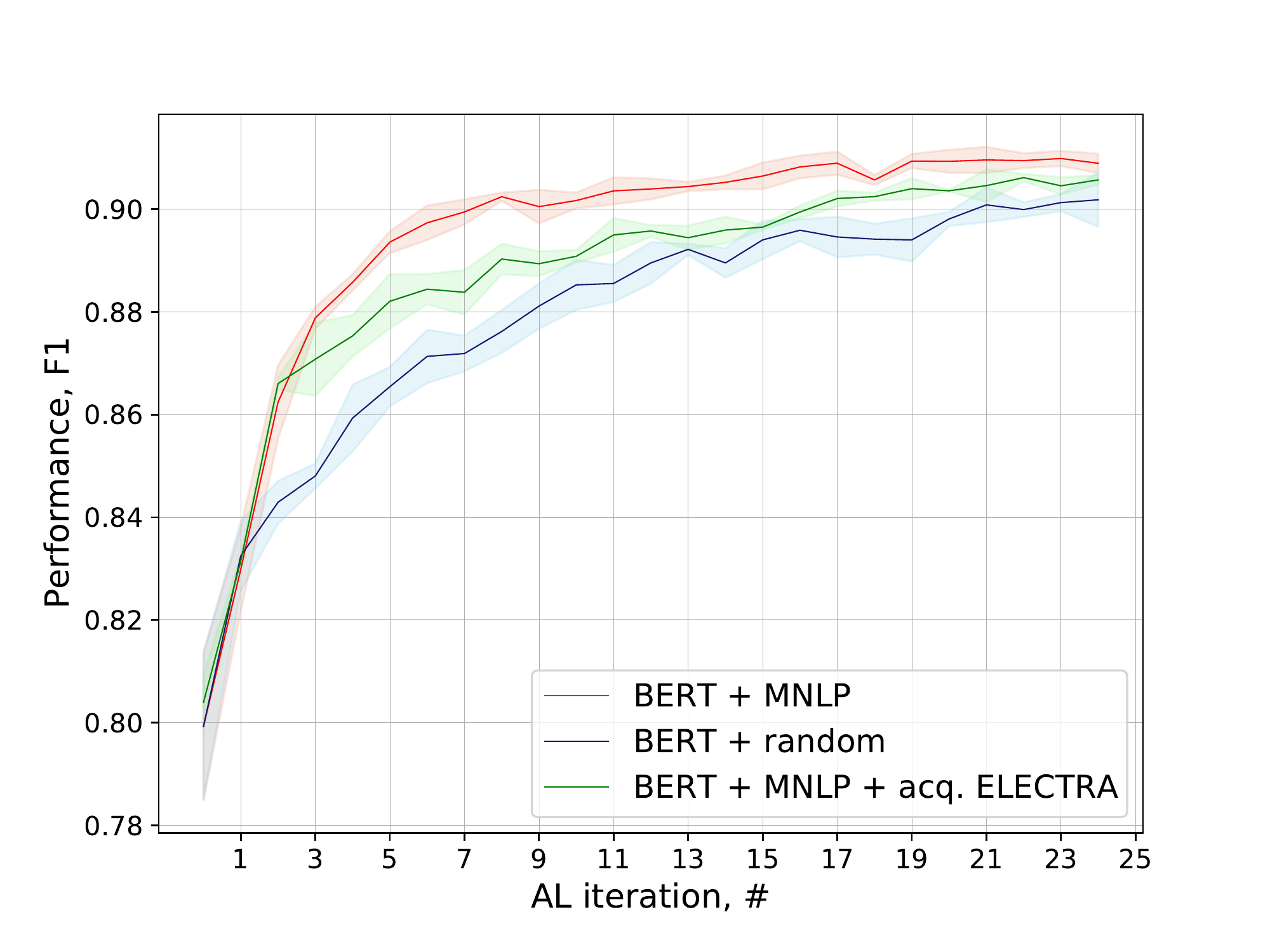} b) ELECTRA is an acquisition model \vspace{0.3cm}}
    \end{minipage}
    
    \caption{AL experiments on the CoNLL-2003 dataset, in which a successor model does not match an acquisition model (for BERT and ELECTRA).}

    \label{fig:bert_electra_acq}
\end{figure*}

\end{document}